\documentclass[conference]{IEEEtran}
\usepackage[]{graphicx}

\usepackage{times}
\usepackage{multirow}

\begin{document}

\title{Context Based Visual Content Verification}

\author{Martin Lukac$^1$, Aigerim Bazarbayeva$^1$, Michitaka Kameyama$^2$\\ 
$^1$ Department of Computer Science\\ Nazarbayev University, Astana, Kazakhstan\\
$^2$ Department of Computer Science\\ Ishinomaki Senshu University, Ishinomaki, Japan
}

\author{\IEEEauthorblockN{Martin Lukac}
\IEEEauthorblockA{Department of Computer Science\\
Nazarbayev University, Astana\\
Kazakhstan}
\and
\IEEEauthorblockN{Aigerim Bazarbayeva}
\IEEEauthorblockA{Department of Computer Science\\
Nazarbayev University, Astana\\
Kazakhstan}
\and
\IEEEauthorblockN{Michitaka Kameyama}
\IEEEauthorblockA{Department of Informatics\\
Ishinomaki Senshu University\\
Ishinomaki, Japan}}

\maketitle

\begin{abstract}
	In this paper the intermediary visual content verification method based on multi-level co-occurrences is studied. The co-occurrence statistics are in general used to determine relational properties between objects based on information collected from data. As such these measures are heavily subject to relative number of occurrences and give only limited amount of accuracy when predicting objects in real world. In order to improve the accuracy of this method in the verification task, we include the context information such as location, type of environment etc. In order to train our model we provide new annotated dataset the Advanced Attribute VOC (AAVOC) that contains additional properties of the image. We show that the usage of context greatly improve the accuracy of verification with up to 16\% improvement. 
\end{abstract}

\IEEEpeerreviewmaketitle

\section{Introduction}

Image or scene understanding is a domain of computer vision aiming to provide a general and detailed description of an image content. The scene understanding goes beyond object detection, recognition, localization and segmentation,  because it aims at providing also the description on a semantic level as well as explaining higher level relations between objects such as behavior or activity~\cite{chella:00,luo:05,sagerer:13}.

The scene understanding can be oriented towards some particular type of scenes or images. For instance Matsuyama~\cite{matsuyama:87} used logic reasoning with inference to improve the understanding of satellite imaging. But with the increasing computing power recently scene understanding has been focusing on general scenes segmentation, semantic segmentation or scene classification~\cite{serrano:04,cao:07,siagian:07,zhang:15}.

With the advent of these more advanced methods and because many of the algorithms are based on machine learning, the quality of the processing heavily depends on the training data and as such there is a strong bias-variance trade-off. Consequently many algorithms resort to higher level content related heuristics such as co-occurrence statistics~\cite{ladicky:10,ladicky:13} to provide a more reliable scene description. 

Content verification~\cite{matsuyama:87,palmeri:04} is a problem related to the analysis of content obtained from sensors and is aimed to insure the quality of result of particular algorithm. It is most commonly used in areas that deal with real-world information such as computer vision, natural language processing or robotic application. Additionally with the increasing complexity of tasks intended to be performed by intelligent agents as well as required reliability, it is necessary to ensure that the perceived information and executed actions are performed correctly. 

In this paper we extend a previously proposed method for content verification in~\cite{lukac:13d,lukac:15,lukac:16b} by enabling a context aware verification approach. While initially the verification method based on advanced co-occurrence statistics was developed for general set of images, we show that restricting context and algorithm can effectively increase the accuracy of verification process. 

This paper is organized as follows. Section~\ref{sec:pre} describes the previous work, in Section~\ref{sec:met} introduces the proposed method of verification and Section~\ref{sec:exp} describes the experiments. Section~\ref{sec:con} concludes this paper. 




\section{Previous Work}
\label{sec:pre}

\begin{figure}[bht]
\centering
	\includegraphics[width=\linewidth]{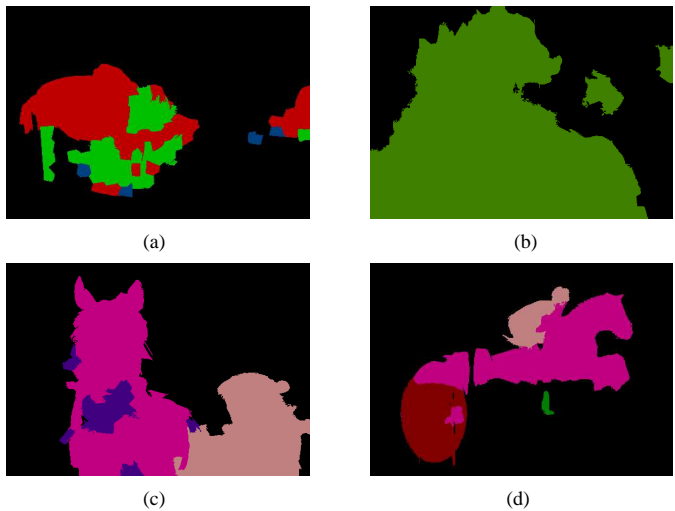}
	\caption{\label{fig:verifex} Example of semantic segmentation results that can be improved by high-level semantic verification. }
\end{figure}

Content verification is required when the result of processing is not guaranteed to be accurate for all possible input scenarios. This is often the case in machine learning when the bias-variance is high due to training set data restriction. In such case the result of image processing for instance, must be verified for inconsistencies resulting from the imperfect processing, lack of information or due to complexity of the task.  For instance, Figure~\ref{fig:verifex} shows a set of examples of semantic segmentation where a high level semantic verification process can solve problems arising from the spurious results. 
\begin{itemize}
	\item Figure~\ref{fig:verifex}(a): parts of sofa (green) are mixed with parts of chair (red) and such result can be semantically detected. 
	\item Figure~\ref{fig:verifex}(b): variations of sizes between several objects of the same class can be detected by relative size comparison
	\item Figure~\ref{fig:verifex}(c): the horse (pink) cannot be a dog (purple) at the same time
	\item Figure~\ref{fig:verifex}(d): a horse (pink) cannot be a cat (dark red) at a same time and a cat  is also rarely under a horse. 
\end{itemize}
Verification can be seen as a part of information restoration process in signal processing or circuit computing~\cite{moon:05}. On a higher semantic level Natural Language Processing (NLP) often repairs the translated or generated content by co-occurrence metrics~\cite{bordag:08,baroni:10,hai:11,levy:14}. NLP uses co-occurrence matrices to determine and to fill in correct words in a sentence or to determine relations between words and a context.

In computer vision reasoning is used to determine the correctness of the image content by analyzing higher-dimensional relations between the obtained content. For instance~\cite{galleguillos:08} uses co-occurrence statistics for object classification, in~\cite{fablet:03} similar approach is used to reason about the geometrical disposition to reason about missing elements using temporal and spatial co-occurrences. In~\cite{lukac:13d,lukac:15} the reasoning is used to remove or modify the existing elements of  semantic segmentation in a meta-learning framework.

A common approach how to improve the accuracy of object recognition and segmentation is to use co-occurrence statistics~\cite{ladicky:10}. Originally used to analyze language~\cite{kroeger:05}, the co-occurrence statistics can appear as a probabilistic generalization of the association analysis rule learning~\cite{piatetsky:91,agrawal:93}.

Co-occurrence statistics represent information collected from sample data and represents probabilistic information about pairs of objects occurring together in one image. It was used on several occasions to improve the recognition accuracy in semantic segmentation and object recognition~\cite{ladicky:10,lukac:13d}. In particular it was used in combination with Conditional Random Fields (CRF)~\cite{ladicky:10}, graph-cut~\cite{ladicky:13} and with Algorithm Selection approach~\cite{lukac:15,lukac:16}. 

In most cases the co-occurrence is used only as existential verifier, i.e. it represents only the statistical information of objects occurring at the same time. More advanced statistics have been proposed in~\cite{lukac:15} where the co-occurrence statistics were generalized to determine statistics of relative size, proximity, position and shape. 

The main problem of co-occurrence statistics is the probabilistic bias that arises when calculating co-occurrence statistics from a data set; i.e. objects occurring more often will tend to bear stronger probabilistic presence than those that occur less often. This becomes a problem if the co-occurrence model or any statistics are used for generative purposes such as used in~\cite{lukac:15}. 

The advantage of the co-occurrence methods is however the relative easiness of using them and creating them from data. Consequently co-occurrence statistics are very useful and can be used both for verification as well as for model generation.

\section{Proposed Method}
\label{sec:met}
In this paper we use the work from~\cite{lukac:13d} as a starting point of our study. The platform introduced in~\cite{lukac:13d} is based on meta-learning and uses algorithm selection to provide optimal output to a computer vision task. As a part of feedback to the algorithm selection mechanism, the content is verified using higher-level relational co-occurrence matrices.

The platform was used in~\cite{lukac:16} and outperformed by intelligent algorithm selection all of the used state-of-the-art semantic segmentation algorithms. Such result was obtained despite the fact that the used semantic verification had a accuracy of 60\% for the 20 categories of objects available in the VOC2012 dataset. 

\begin{figure}[bht]
	\centering
	\includegraphics[width=0.7\linewidth]{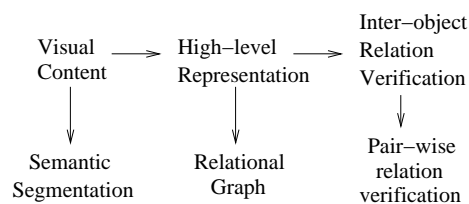}
	\caption{\label{fig:ver1} The process of semantic verification using object relationships.}
\end{figure}

The general scheme of the verification method used in~\cite{lukac:16}  is shown in Figure~\ref{fig:ver1}. 

The verification uses the result of semantic segmentation such as shown in Figure~\ref{fig:verifex}, builds a relational model (called high level representation in Figure~\ref{fig:ver1}) and determines if these relations are semantically valid or lead to a contradiction. 

For instance a human will never be in the air without the support unless the context is specific to sky diving or jumping in the air. 

\begin{figure}[bht]
\centering
	\includegraphics[width=\linewidth]{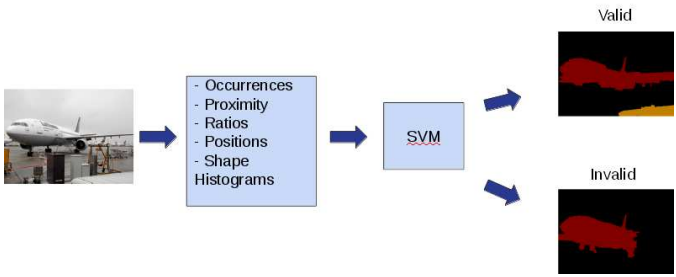}
	\caption{\label{fig:fbv}Co-occurrences based Verification schematic of example}
\end{figure}

The main principle of this verification process is co-occurrence statistics of four different kinds. These co-occurrence statistics are used to determine the properties between any two objects obtained as a result of semantic segmentation. The collected information about objects were represented as a vector for each pair of objects in the image. Then from the set of all relationships a general conclusion on the object configuration is obtained by majority voting or by SVM. In the work used in this paper an SVM was trained to discriminate whether yes or no a contradiction exists in the semantic segmentation (Figure~\ref{fig:fbv}). 

Unlike in simple co-occurrence statistics, the original work from~\cite{lukac:15}  uses four distinct measures to estimate the contradiction. The schematic representation of the original method is shown in Figure~\ref{fig:fbv}. Unlike the work in~\cite{lukac:16} however in~\cite{lukac:15}  the accuracy of semantic verification was up to 92\% of accuracy but on a significantly smaller dataset and with specially prepared data. Consequently the semantic verification based on co-occurrence statistics should be studied in more depth to understand better the capacity to capture and represents real world information.

\begin{figure}[ht]
	\centering
	\includegraphics[width=0.6\linewidth]{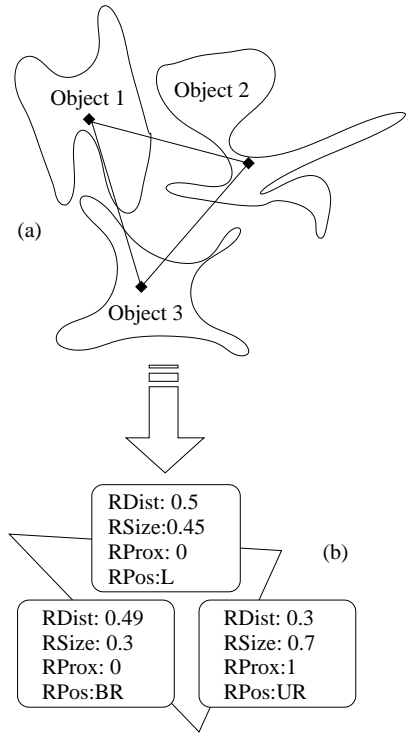}
	\caption{\label{fig:coocs}Example of co-occurrence information extracted from a set of neighboring objects}
\end{figure}

The five features obtained form semantic segmentation are described as follows:
\begin{enumerate}
	\item The first is simple co-occurrence collected from the whole training data set. The result was a set of coefficient representing the probability that two objects occur at the same time in one image. 
	\item The second is the position statistics. For this to work each object in the image was segmented and center of gravity was calculated. For each two pairs of objects eight possible vector orientations were available. These are shown in Figure~\ref{fig:coocs}.
	\item The third co-occurrence was calculated using proximity of objects. The coefficient indicating representing for any two objects the probability that the two objects are in visual contact. In fact in~\cite{lukac:16} the proximity was integrated and was given as additional four co-occurrence statistics: on, under, front and back.
	\item The fourth co-occurrence is obtained from sizes of objects. Similarly to the previous statistics, from the segmented object the number of pixels is used as size. The normalized average of the size ratio indicates the probability of two objects relative sizes. 
\end{enumerate}
A shape characterization was added to increase the probability of recognizing the segmented object with higher accuracy. For this the shape histogram approach is used. The shape histogram is simply created by sampling the boundary of recognized object and clustering the distance of each sampled object from the center of gravity. 

An example image and a high level representation is shown in Figure~\ref{fig:coocs}. Figure~\ref{fig:coocs}(a) shows three segmented objects in their relative position. The objects' gravity centers being the vertices of a triangle are joined by the edges representing pairwise relations between objects. The relations are represented by the four relative co-occurrences relative distance (RDist), relative size (RSize), relative proximity (RproX) and relative position (RPos) as shown in Figure~\ref{fig:coocs}(b). The relative position is ordered from left-to-right so that relative position L (left) indicates the position of the left object with respect to the right one and so on. 

One of the main reasons of the low accuracy of the co-occurrence statistics used for semantic verification is due to statistical co-occurrence bias occurring because of lack of training data with same number of objects of different classes. For instance, because most of the images are human centric, human is present in most of the images and thus all the statistics are biased toward an over-presence of human. 

In order to avoid this statistical co-occurrence based bias, the proposed approach formulates $n$ models of contextually \em tinted \em models. The concept is simply explained as follows: a co-occurrence statistics will depend on the context that defines probability of certain objects occurring at all or in some particular configuration.

\subsection{Context Aware Verification}

\begin{figure}[bht]
\centering
	\includegraphics[width=\linewidth]{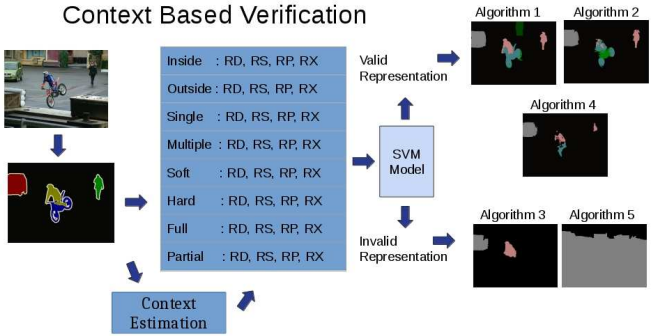}
	\caption{\label{fig:cbv}Co-occurrences based Verification schematic of example}
\end{figure}

Considering the context, the environment and the general specification of the application the verification process can be broken from a single general verification algorithm to a larger number of smaller verifiers. This idea is shown in Figure~\ref{fig:cbv} where several context-aware verification models are used for each of the existing defined contexts. 

Figure~\ref{fig:cbv} shows that for any input image, from the existing semantic segmentation or provided by user a context is estimated. The context is then used to select more accurate and smaller version of the general verifier.

The idea of the proposed improvement in this paper is to break the problem into categories or environments that allows to reduce efficiently the size of the co-occurrence matrices by eliminating objects that do not occur together or not at all. 

For instance, from the VOC2012 data it was observed that if there is a person on a horse it is always outside, while the presence of cat is exclusively limited to inside environment. 

Naturally such observation is highly depending on the dataset but it is reasonable to assume that the database images represent real objects in context proportionally to their natural existence in real world. For instance, domesticated cats are mostly inside while outside their occur much less often.   

\subsection{Selecting Contexts}

Using the context characterization it is necessary to determine meaningful contexts that would allow improved verification and simplified reasoning. For this we generated a new set of visually relevant attributes for the VOC2012 data~\cite{everingham:10} set called the Advanced Attributes VOC (AAVOC). There exists already various augmented image sets by attribute however in most of the cases these attributes are object related or a particular view limited. Additionally the attributes are mostly and only focused on the image primary objects rather to the holistic properties of the environment. Consequently a new data set was created for this purpose solving the two above mentioned problems. 

The used dataset provides thus the following improvements:
\begin{enumerate}
	\item Object attributes labeling: only predefined attributes are considered as being or not being present
	\item Scene attributes relating to global scene properties such as background or environment
	\item Aperture attributes such as exposure quality or type of photo shot
\end{enumerate}

A total of 44 attributes have been used. Out of these twelve are used to specify category of a given object and thus were not used in the candidates of global contexts. The remaining 32 attributes have together 172 different values. Consequently the provided labelling can be seen as having 172 binary attributes. 

From each image in the AAVOC data set, each object was extracted sequentially and the attributes were provided. Thus according to the scopes of the attribute, for one image the value of particular attribute can remain the same or change.

For each of the available labels, first the empty values have been replaced by a placeholder. Then the entropy for each attribute and the labels was calculated using the Shannon's entropy the mutual information according to $I(L,A)= \sum_{l\in L}\sum_{a\in A}p(l,a)log\left ( \frac{p(l,a)}{p(l)p(a)}\right )$ where $p(l)$ and $p(a)$ are prior probabilities of $l\in L$ and $a\in A$.  

The attributes we are looking for are such context variables that best describes the data from a general point of view. The attribute of highest interest should have thus the following properties:
\begin{enumerate}
	\item It has a defined value in all images
	\item Values of the attribute are proportional to all images
	\item Allows to estimate the classes of objects most reliably
\end{enumerate}

For instance, environment location inside or outside is example of such attribute that satisfied most of the criteria. Specifically, any attribute specific to any object is in most cases unsuitable because it is specific to either a type of object, a group or a description. 

The main rule for selecting such attribute is also one that has defined values for all images and one that allows to best predict and separate between classes of objects. For instance, an attribute separating objects in two groups of non overlapping sets is desired. Attribute \em inside \em will indeed put all sky in one group (being not \em inside\em) and all TV monitors in another group 8being \em inside\em).

\section{Experimentation}
\label{sec:exp}

To evaluate the proposed method the VOC2012 challenge dataset was used. Two subsets were used. The \em train \em subset was used to build our models and the \em val \em dataset was used to evaluate the proposed approach. 

The verification framework is the one used in~\cite{lukac:15}. In the proposed approach verification is required in order to verify the semantic content of each semantic segmentation. The method evaluation and accuracy of predicting semantic contradiction  from a segmentation results are shown in Table~\ref{tab:verif0} on the VOC 2012 validation dataset. 

\begin{table}[bht]
\centering
	\caption{\label{tab:verif0} Verification of semantic segmentation accuracy result using the method from~\cite{lukac:15}}
	\begin{tabular}{|cc|c|}
		\hline
		\multirow{2}{*}{Algorithm}& & Verification\\
		&  & Accuracy\\
		\hline
		~\cite{simonyan:15}& & 79\%\\
		~\cite{chen:14}& & 75\%\\
		~\cite{bharath:14}& & 74\%\\
		~\cite{ladicky:13}& & 67\%\\
		~\cite{ion:11}& & 39\%\\
		\hline
		Average& &66.8\%\\
		\hline
	\end{tabular}
\end{table}
The accuracy of semantic segmentation represents a pixel-wise difference between a ground truth semantic segmentation and the algorithm's result. An accuracy of 50\% can represent several different results: (a) exactly 50\% of objects have been perfectly segmented only and all others have not been even detected, (b) the average accuracy of detecting and segmenting images is 50\% and (c) all objects were properly detected and have been segmented with 50\% of accuracy.

The original algorithm tested in~\cite{lukac:15,lukac:16} had an accuracy of detecting a contradiction of 60\%. The algorithm was trained and tested according to the description given in ~\cite{lukac:15}. 

The experiments using the context based verification were separated in two sets. First, an algorithm and context dependent verification algorithm is implemented and was evaluated on the validation data set of the VOC2012 dataset. The idea behind this approach to semantic verification is to apply fine grain verification able to trained for each available algorithm individually. Such approach, could precisely detect semantic segmentations contradictions resulting from the particular structure and architecture of each algorithm.

In the second set of experiments, a more general approach is used. Verification algorithms have been constructed only for each of the available image contexts.

\begin{figure}[bht]
	\centering
	\includegraphics[width=\linewidth]{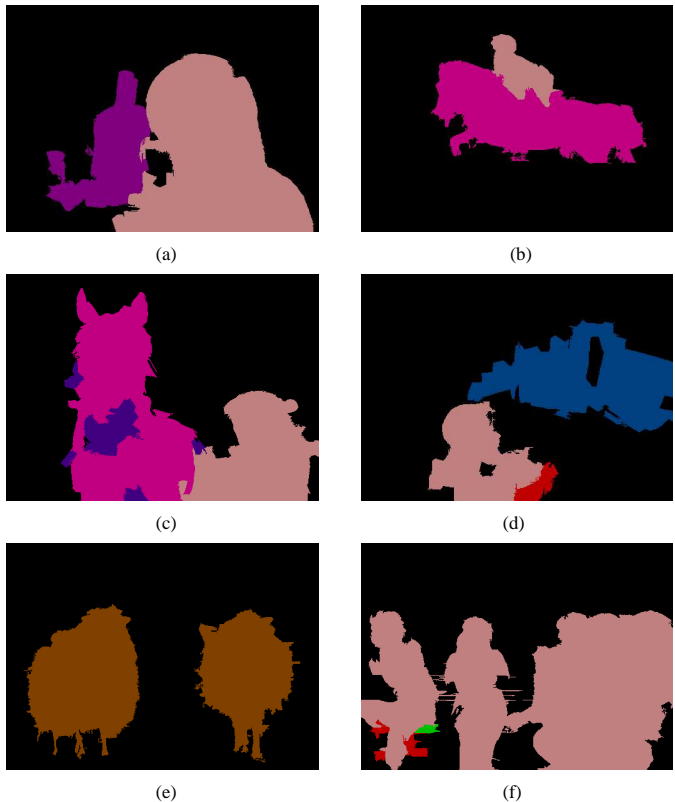}
	\caption{\label{fig:excex} Example of (a-b) valid images containing no contradiction, (c-d) valid images containing contradiction and (e-f) invalid images for the verification algorithms}
\end{figure}

In order to evaluate the method on the simplest available cases of scene configuration only images that contained single instances of class were used for both training and testing. Example and counter example of such images are shown in Figure~\ref{fig:excex}.
Following the approach in~\cite{lukac:15} the used data to train the verification algorithms were obtained from the tested algorithms output. The valid images in Figure~\ref{fig:excex} represent the negative examples for the training of the contradiction detector. Positive examples are generated by removing one of the available objects.

Table~\ref{tab:verif} shows the results and accuracy of a context and algorithm dependent verification. Each row represents one of the selected contexts and each column represents one particular algorithm. 

\begin{table}[bht]
	\centering
	\caption{\label{tab:verif} Content verification algorithms for each algorithm and for each context.}
	\begin{tabular}{|c|c|c|c|c|c|c|}
		\hline
		Context / Dataset& ~\cite{chen:14}&~\cite{ion:11}&~\cite{ladicky:13}&~\cite{bharath:14}&~\cite{simonyan:15}&AVG\\
		\hline
		outside&80 \%&58 \%&62 \%&77 \%&81 \%&71.6\%\\
		inside&69 \%&76 \%&72 \%&76 \%&76 \%&73.8\%\\
		single&75 \%&63 \%&66 \%&81 \%&78 \%&72.6\%\\
		multiple&81 \%&50 \%&75 \%&75 \%&50 \%&66.2\%\\
		soft&72 \%&57 \%&62 \%&82 \%&79 \%&70.4\%\\
		hard&73 \%&67 \%&62 \%&85 \%&76 \%&72.6\%\\
		full&75 \%&63 \%&67 \%&79 \%&80 \%&72.8\%\\
		partial&74 \%&55 \%&62 \%&76 \%&76 \%&68.6\%\\
		\hline
		Average&75 \%&61 \%&66 \%&79 \%&75 \%&71.2\%/71.07\%\\
		\hline
	\end{tabular}
\end{table}

The results in the Table~\ref{tab:verif} shows that certain algorithms are more consistent and thus more predictable than others. In particular, three out of the five tested algorithm were convolutional neural networks (CNN)~\cite{bharath:14,chen:14,simonyan:15}  while the two remaining ones are older algorithms based on Conditional Random Fields (CRF)~\cite{ladicky:13} and on machine learning of region segmentation~\cite{ion:11}.

Interestingly, the accuracy of determining if an algorithm's semantic segmentation contains a semantic contradiction is proportional to the average accuracy of the algorithm itself. This indicates that the used contradiction detection method principles are not capturing accurately and efficiently the algorithm-dependent error structure. 

To verify this claim another set of experiments was conducted. This set of experiments was only context dependent and the contradiction detection method was applied on the semantic segmentation of all used algorithms. The results are shown in Table~\ref{tab:verifall}. Column one indicates the context, column 2, 3 and 4 indicates the number of images used to train the verifier, column five indicates the obtained accuracy of predicting a contradiction, column six shows the results reported in Table~\ref{tab:verif} and column seven shows the relative improvement. 
\begin{table}[bht]
	\centering
	\caption{\label{tab:verifall} Content verification for each content but algorithm independent.}
	\scalebox{0.8}{
	\begin{tabular}{|c|c|c|c|c|c|c|}
		\hline
		\multirow{3}{*}{Context}& \multirow{3}{*}{Valid}&\multirow{3}{*}{Invalid}&\multirow{3}{*}{Total}&Result of&Average of& \multirow{3}{*}{Improvement}\\
		& & & & verification &previous& \\
		& & & & on all datasets&results& \\
		\hline
		inside\_all&1732&739&2471&74.00\%&73.80\%&0.20\%\\
		outside\_all&3301&793&4094&76.00\%&71.60\%&4.40\%\\
		multiple\_all&1537&474&2011&68.00\%&66.20\%&1.80\%\\
		single\_all&4241&1300&5541&75.00\%&72.60\%&2.40\%\\
		partial\_all&2872&913&3785&72.00\%&68.60\%&3.40\%\\
		full\_all&3448&1023&4471&76.30\%&72.80\%&3.50\%\\
		soft\_all&3168&1110&4278&73.00\%&70.40\%&2.60\%\\
		hard\_all&2806&683&3489&73.00\%&72.60\%&0.40\%\\
		\hline
		Average&--&--&--&73.41\%&71.07\%&--\\
		\hline
	\end{tabular}}
\end{table}
The results from this second set of experiments are shown in Table~\ref{tab:verifall}. Interestingly the results show that indeed he verification is same or even performs better when performed in a algorithm independent manner rather than when done on an algorithm by algorithm basis. This implies that if there is an underlying structure to each algorithm's error then either it is negligible with respect to the context or it cannot be properly captured by the model used here.

The general remarks on the results include the observation that while the improvement of the verification accuracy compared to the initial result from~\cite{lukac:16} is considerable the accuracy claimed in~\cite{lukac:15} was not achievable in any of the here performed experiments. The conclusion from this observation is that while in~\cite{lukac:15} the train and test data was the problem, in~\cite{lukac:16} the problem was clearly the fact that single verifier for all algorithms and all contexts is too complex and inaccurate. 

Surprisingly splitting the verifier to a set of smaller ones w.r.t. context and algorithms perform worse than only train verifiers one for each available contexts. This is very interesting because this indicates that the type of error obtained as a result of semantic segmentation is strongly dependent on the context of the image rather than on the type of the used algorithms. 

\section{Conclusion}
\label{sec:con}
In this paper, we showed a context based approach to semantic segmentation content verification. We experimentally demonstrated that the context is an important variable allowing to strongly increase the accuracy of the verification while a finer granularity of verification (algorithm dependent verification) did not improve the accuracy. 

Additionally we have experimentally shown that the most important factor in increasing the accuracy of content verification is indeed the context. Thus in the future work more development in the area of automatic context extraction and identification is to be studied and developed.

\bibliographystyle{IEEEtran}
\bibliography{./bibexport}

\end{document}